\documentclass[cameraready]{Interspeech}

\newcommand\singleSENSE{Att(sem)}
\newcommand\multiSENSE{Att(sem+spk)}
\newcommand\singleSENSEspk{Att(spk)}

\title{Learning Multiple Utterance-Level Attribute Representations with a Unified Speech Encoder}

\author[orcid=0000-0000-0000-0000]{Maryem}{Bouziane}
\author[orcid=0000-0002-8472-6890,  correspondingauthor]{Salima}{Mdhaffar}
\author[orcid=0000-0002-3656-8883, correspondingauthor]{Yannick}{Estève}
\address{
    LIA - Avignon Université, France
}

\email{firstname.lastname@univ-avignon.fr}


\usepackage{comment}

\begin{document}

\maketitle

\begin{abstract}
Speech foundation models trained with self-supervised learning produce generic speech representations that support a wide range of speech processing tasks. When further adapted with supervised learning, these models can achieve strong performance on specific downstream tasks. Recent post-training approaches, such as SAMU-XSLR and SONAR, align speech representations with utterance-level semantic representations, enabling effective multimodal (speech–text) and multilingual applications. While speech foundation models typically learn contextual embeddings at the acoustic frame level, these methods learn representations at the utterance level.

In this work, we extend this paradigm to arbitrary utterance-level attributes and propose a unified post-training framework that enables a single speech foundation model to generate multiple types of utterance-level representations. We demonstrate the effectiveness of this approach by jointly learning semantic and speaker representations and evaluating them on multilingual speech retrieval and speaker recognition tasks.
\end{abstract}

%%%% le plan :

\section{Introduction}

Recent advances in self-supervised speech representation learning have led to the emergence of large speech foundation models capable of capturing rich acoustic and linguistic information from raw audio. 
Models such as wav2vec~2.0~\cite{baevski2020wav2vec}, HuBERT~\cite{hsu2021hubert}, and w2v-BERT~\cite{chung2021w2v} are trained on massive amounts of unlabeled speech and learn contextualized representations that can be adapted to a wide range of downstream tasks \cite{parcollet2024lebenchmark, yang2021superb}. 
These models typically produce frame-level acoustic representations, which are highly effective for tasks such as automatic speech recognition or speech translation.
Beyond frame-level representations, there has been growing interest in learning utterance-level speech embeddings that capture higher-level information such as semantics or speaker traits. 
Such representations are particularly useful for tasks including speech retrieval, cross-modal search, speaker verification, or conversational understanding. 
Recent post-training approaches have demonstrated that speech encoders can be aligned with text-based semantic embedding spaces, enabling multilingual and multimodal applications such as speech–text retrieval and speech translation search. 

A notable example is the SENSE framework~\cite{mdhaffar2025sense}, close to the Meta's SONAR framework~\cite{Duquenne:2023:sonar_arxiv}, which learns semantic speech embeddings through a teacher–student knowledge distillation paradigm~\cite{hinton2015distilling}: in this approach, a pretrained text embedding model provides semantic targets, while a speech encoder is trained to map speech utterances into the same semantic space. 
This alignment allows speech representations to directly capture the meaning of an utterance, independent of language, making them suitable for multilingual semantic retrieval tasks~\cite{mdhaffar2025sense,Duquenne:2023:sonar_arxiv,khurana2022samu}. 

However, aligning speech representations exclusively with semantic embeddings introduces an important limitation: paralinguistic information present in speech may be suppressed. 
In particular, attributes such as speaker identity, emotion, or speaking style are not preserved when the representation is optimized solely to match textual semantic embeddings. 
This raises an important question: can a single speech encoder learn representations that simultaneously capture multiple utterance-level attributes?

In this work, we propose a unified post-training framework that allows a single speech foundation model to produce multiple utterance-level representations corresponding to different attributes. 
Our approach extends the teacher–student paradigm by introducing multiple task-specific supervision signals, each defining a target embedding space for a particular attribute. 
A shared speech encoder is jointly trained to align with these different targets through task-specific projection branches.
To demonstrate the effectiveness of this framework, we focus on learning two complementary utterance-level attributes: semantic representations, obtained through alignment with multilingual text embeddings, and speaker representations, obtained through supervision from a pretrained speaker verification model.

We evaluate the proposed model on two representative tasks. 
First, we measure the quality of the semantic representation using multilingual speech and speech–text translation retrieval benchmarks. 
Second, we evaluate the speaker representation on the VoxCeleb speaker verification task. 

In this paper, we make the following contributions:
\begin{enumerate}
    \item We introduce a general multi-task teacher–student framework for learning multiple utterance-level attribute representations from a shared speech encoder.
    \item We demonstrate that semantic and speaker representations can be learned jointly without significantly degrading the performance of either attribute.
    \item We provide an analysis of layer usage across tasks, showing how semantic and speaker information are distributed differently within the shared encoder.
\end{enumerate}

\section{Method}

Our work extends the teacher-student distillation paradigm used in the SENSE framework~\cite{mdhaffar2025sense}, an open-source solution designed for learning utterance-level semantic speech representations. 
The SENSE framework is derived from SAMU-XSLR~\cite{khurana2022samu}, an approach close to the Meta SONAR framework.

\subsection{The SENSE framework}

SENSE operates on a teacher-student knowledge distillation paradigm, where the primary objective is to align speech and text within a shared, language-agnostic semantic space. 
Specifically, the architecture utilizes a pre-trained text embedding model acting as the teacher and a speech encoder acting as the student. The speech encoder is initialized with an SSL speech encoder. 
On the text side, the teacher is a language-agnostic sentence embedding generator, like LaBSE~\cite{feng2022language} or BGE-M3~\cite{chen2024bge}. 

To align speech representations with utterance-level semantic representations, the frame-level representations from the last layer of the speech encoder are first aggregated by an attentive pooling layer. The resulting vector is then passed through a linear projection and an activation function, producing one utterance-level speech representation for each speech segment.

The entire model, made of the initial speech encoder, the attentive pooling layer, and the linear projection, is trained to maximise the cosine similarity between the utterance-level speech vector and the sentence-level text embedding given by the teacher model.
Throughout the training process, the text encoder remains strictly frozen to preserve its semantic structure, whereas the speech encoder is optimized. By directly optimizing the cosine similarity, semantic knowledge is transferred from the text modality to the speech modality, enabling the speech encoder to learn meaning-oriented representations at the utterance level. 

\subsection{Learning multiple utterance-level attribute representations}

We extend the teacher–student paradigm to learn multiple utterance-level representations corresponding to different attributes from a shared speech encoder. 
Let $\mathcal{T}$ denote the set of target attributes and $\tau \in \mathcal{T}$ one particular attribute.
Given an input speech signal, the pretrained SSL speech encoder produces a sequence of hidden representations at different layers:
$
H^{(\ell)} \in \mathbb{R}^{T \times D},
$
where $T$ is the number of time frames, $D$ is the hidden dimension of the encoder representations, and $\ell$ denotes the layer index.

For each attribute $\tau$, a task-specific branch is attached to the shared encoder in order to produce an utterance-level embedding aligned with the corresponding teacher representation.
For each selected encoder layer representation $H^{(\ell)}$,
a linear projection specific to attribute $\tau$ is applied:
$\tilde{H}^{(\ell)}_{\tau}
=
H^{(\ell)} W^{(\ell)\top}_{\tau} + b^{(\ell)}_{\tau}.
$

This projection maps the shared encoder representations into a feature space associated with attribute $\tau$. 
The purpose of this transformation is to limit the amount of adaptation required from the shared speech encoder. 
Instead of forcing the encoder to directly produce representations suitable for all target attributes, the model learns attribute-specific projections that transform the shared representation space into the embedding space required by each attribute.
In this way, the speech encoder can preserve a general-purpose representation of the input signal, while each attribute branch adapts this shared representation to its own target space. 
This separation helps the encoder to remain task-agnostic and reduces potential interference between attributes when the model is trained with multiple supervision signals.
This transformation maps the encoder representations into a task-specific feature space associated with attribute $\tau$.

Different attributes may rely on different regions of the encoder. 
To capture this behavior, the model learns a scalar importance score $s_{\tau,\ell}$ for each layer. 
This layer-weighting mechanism is not present in the SENSE framework.
These scores are converted into normalized interpolation weights using a softmax function:
$
\lambda_{\tau,\ell}
=
\frac{\exp(s_{\tau,\ell})}
{\sum_{j=1}^{n} \exp(s_{\tau,j})},
\qquad
\sum_{\ell=1}^{n} \lambda_{\tau,\ell} = 1.
\label{eq:layer_weights}
$

The projected representations are then combined through a weighted sum:
$
\hat{Z}_{\tau}
=
\sum_{\ell=1}^{n}
\lambda_{\tau,\ell} \tilde{H}^{(\ell)}_{\tau}
$.

A layer normalization is then applied: $Z_{\tau} = \mathrm{LayerNorm}(\hat{Z}_{\tau}).$
The frame-level sequence $Z_{\tau}$ is aggregated into a single utterance-level representation using an using an attribute-specific attention pooling mechanism: $p_{\tau} = \mathrm{AttentionPooling_\tau}(Z_{\tau})$.

Depending on the target representation space associated with attribute $\tau$, an optional linear projection can be applied.
%\begin{equation}
%q_{\tau} = P_{\tau} p_{\tau} + c_{\tau},
%\end{equation}
%where $P_{\tau}$ and $c_{\tau}$ define an attribute-specific mapping into the target embedding space.

\begin{figure}[h]
  \centering
  \includegraphics[width=0.7\linewidth]{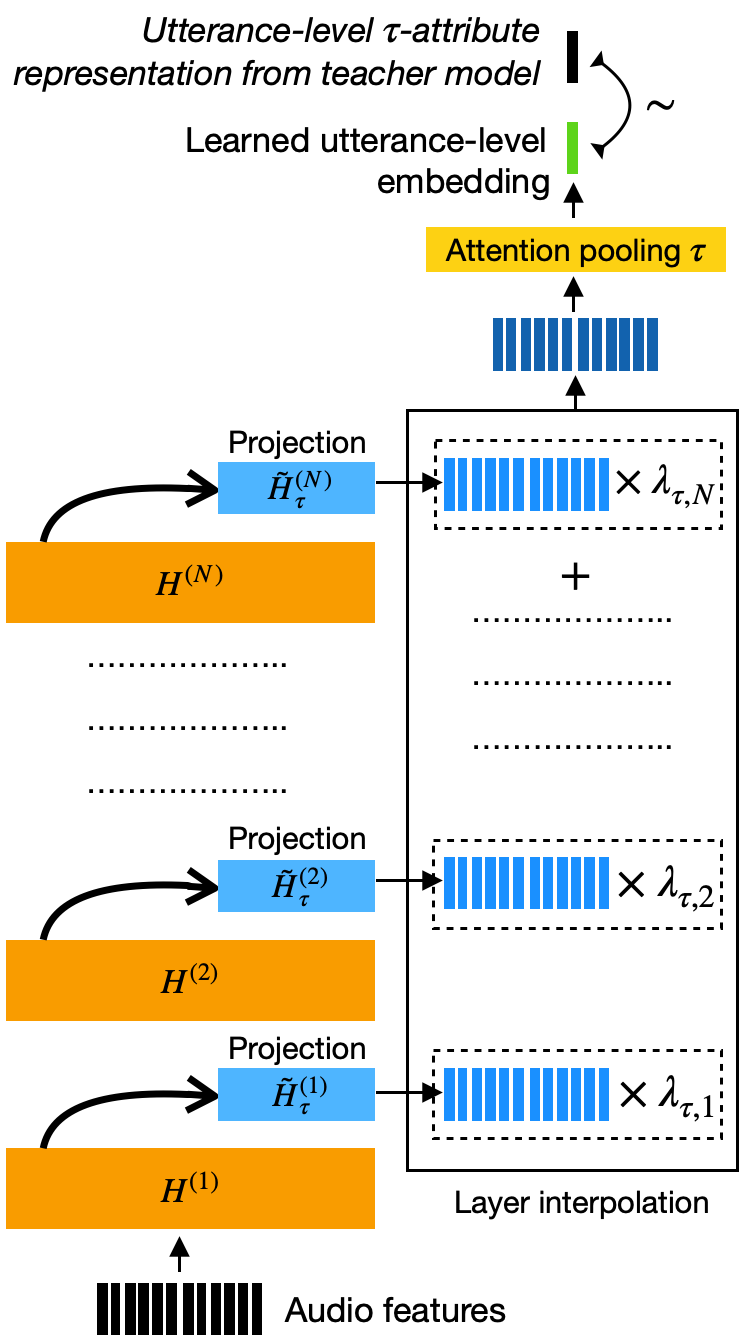}
  \caption{Attribute-specific branch used to learn an utterance-level representation for attribute $\tau$ in the teacher–student framework. Layer representations from the shared SSL encoder are projected, combined using learnable interpolation weights $\lambda_{\tau,\ell}$, and aggregated by attention pooling to produce an embedding aligned with the teacher representation.}
  %\caption{Branch architecture to learn a utterance-level representation for a target attribute $\tau$ in a teacher-student framework. The orange blocks correspond to the pretrained SSL encoder. 
  %For each target attribute, layer-wise linear projections, learnable layer-interpolation weights, and an attention pooling layer are added on top of the shared encoder to produce an utterance-level embedding aligned with the corresponding teacher representation. In a multi-task setting, this branch is instantiated for each attribute.}
  \label{fig:learning_arch}
\end{figure}

The final utterance-level embedding produced by attribute branch $\tau$ is $\ell_2$-normalized and then aligned with the corresponding teacher embedding using a cosine similarity objective.
Figure~\ref{fig:learning_arch} summarizes this approach.

The model is trained using a multi-task learning framework in which each target attribute is considered as an individual task. 
For each attribute $\tau$, a dedicated branch is instantiated. 
All parameters, including those of the SSL encoder, are jointly optimized during training.

\section{Experiments}
\subsection{Experimental setup}

We target two tasks in our multi-task teacher-student learning: (1) a semantic-based task, focusing on learning language-agnostic semantic content representations, and (2) a speaker-based task, aiming to characterize the speaker information within the same unified architecture.
The semantic branch is trained to align the utterance-level speech representation with the corresponding BGE-M3~\cite{chen2024bge} semantic embedding, while the speaker branch is trained to align the utterance-level speech representation with a trained ECAPA-TDNN~\cite{desplanques2020ecapa} speaker embedding model. 
The model ECAPA-TDNN\footnote{https://huggingface.co/speechbrain/spkrec-ecapa-voxceleb} used in this work is trained with VoxCeleb 1~\cite{nagrani2020voxceleb} and  VoxCeleb 2~\cite{chung2018voxceleb2} datasets.
Both teacher models remain frozen throughout training.
The speech encoder model is initialized with w2v-BERT 2.0~\cite{barrault2023seamless}.
We implemented our multi-task model in SpeechBrain~\cite{ravanelli2024open} based on the open source SENSE~\cite{mdhaffar2025sense} implementation\footnote{\url{https://github.com/speechbrain/speechbrain/tree/develop/recipes/CommonVoice/SENSE}} and trained it on the Common Voice 19 dataset~\cite{ardila2020common} using the 83 languages supported by BGE-M3, corresponding to 8,250 hours of speech. 
Similar to~\cite{mdhaffar2025sense}, we use only the validated training dataset from Common Voice.
To account for language imbalance, training examples are sampled with a weighted sampling strategy~\cite{khurana2022samu}. 
We fine-tune the shared w2v-BERT 2.0 encoder with Adam using a learning rate of $10^{-5}$. 
The task-specific modules are optimized separately with Adadelta using an initial learning rate of 1.5. Training is performed with a batch size of 20 for both training and validation.
This model is trained for 350K iterations by employing 8 H100 GPUs.
%We use the sum of the semantic and speaker cosine losses as the training objective, with equal weights for both tasks.
\vspace{-0.2cm}
\subsection{Evaluation}
Our primary objective is to assess whether the multi-task model can effectively learn both semantic and speaker representations without degrading the performance of either task compared to single-task training. 
To put its capabilities into perspective, we evaluate the model on both semantic and speaker tasks.
\begin{itemize}
    \item Semantic task: We assess the model’s performance on multilingual and multimodal retrieval. %measuring how well it captures language-agnostic semantic representations across different languages and modalities. 
    %In this setting, spoken utterances in one language must be matched to written or oral translations in another language. 
    \item Speaker task: We evaluate the model on a speaker verification task, which aims to determine whether two audio samples originate from the same speaker.
\end{itemize}

For the semantic evaluation, we compare our multi-task model with two state-of-the-art models, SENSE and SONAR, both designed to learn semantically meaningful speech representations. 
For the speaker verification task, we compare our model with the ECAPA-TDNN speaker embedding model. 
In addition, we train a single-task baseline by retaining only the speaker branch of our architecture. 
This baseline allows us to assess the contribution of the multi-task learning framework and to quantify the impact of joint semantic and speaker supervision on speaker verification performance.

\vspace{-0.2cm}
\subsubsection{Multilingual and multimodal translation retrieval}

To evaluate whether the proposed multi-task training preserves the semantic quality of the learned speech representations, we assess the model on multilingual translation retrieval.
For each experiment, we define a query set and a search set. 
The query set always contains speech utterances, while the search set contains either speech or text in another language, depending on the evaluation condition. 
The objective is to retrieve the correct translation of each query from the candidates in the search set.
Each query utterance is encoded into an utterance-level embedding using the semantic branch of the multi-task model. 
The items in the search set are encoded according to their modality. 
When the search set contains speech, utterance-level speech embeddings are extracted with the same speech encoder. 
When it contains text, each sentence is represented using the corresponding frozen text encoder: BGE-M3 for SENSE and our multi-task model, and the SONAR text encoder for SONAR, based on the NLLB model~\cite{costa2022no}. 
After normalization via mean subtraction, retrieval is performed by comparing each query embedding with all candidate embeddings in the search set using cosine similarity. 
Performance is reported with Recall@1.
We consider the following evaluation settings. 
For (1) speech $\rightarrow$ speech retrieval, we use the VoxPopuli~\cite{wang2021voxpopuli} dataset, where both the query set and the search set contain speech utterances in different languages.  
    %We evaluate three directions: from a non-English language to English, from English to another language, and between two non-English languages. 
For (2) speech $\rightarrow$ text retrieval, we use the MTEDx~\cite{salesky21_interspeech} where the query set contains speech utterances and the search set contains the corresponding translated text sentences in another language.
 In order to evaluate the generalization ability of the models to
unseen and low-resource languages, we additionally used the FLEURS dataset~\cite{conneau2023fleurs}.

\vspace{-0.2cm}
\subsubsection{Speaker verification}
We evaluate the quality of speaker-related information through a speaker verification experiment using the VoxCeleb1-O evaluation protocol. 
Speaker embeddings are extracted from both enrolment and test utterances using the speaker branches of our models (trained either in single-task or multi-task mode) or using representations obtained from the ECAPA-TDNN model. 
Verification scores for each trial pair are computed using cosine similarity. 
Performance is reported in terms of Equal Error Rate (EER) and minimum normalized Detection Cost Function (MinDCF) with $P_{\text{target}} = 0.01$ and $C_{\text{FA}} = C_{\text{Miss}} = 1$.

\subsection{Results}

This section presents the experimental results for the semantic and speaker evaluation tasks described above. 
In the result tables, we use the following notation for clarity: SONAR refers to the Meta AI SONAR model introduced in~\cite{barrault2023seamless}; Att(sem) corresponds to the SENSE model trained with a semantic objective as described in~\cite{mdhaffar2025sense}; Att(spk) denotes the single-task variant of our architecture trained only with the speaker objective; and Att(sem+spk) represents the proposed multi-task model trained jointly with semantic and speaker supervision.

%Overall, the proposed multi-task model learns both attributes effectively: it remains very close to the semantic-only SENSE baseline on translation retrieval, while also producing speaker representations that are highly consistent with the ECAPA-TDNN teacher.

\vspace{-0.2cm}
\subsubsection{Multilingual and multimodal translation retrieval}

Table~\ref{tab:Speech_speecH_retr} presents R@1 scores for speech-to-speech translation retrieval on VoxPopuli for the three models SONAR, \singleSENSE\ and \multiSENSE. \\
\singleSENSE\ and \multiSENSE\ use a single multilingual speech encoder shared across all languages. SONAR, on the other hand, uses 37 language-specific encoders. 
In our experiments, we use the corresponding SONAR encoder when available, and the English encoder otherwise.
Across the evaluated language pairs, our multi-task model remains very close to the \singleSENSE, with only small differences in R@1, and consistently outperforms SONAR. 
This shows that adding speaker supervision preserves the semantic retrieval ability to a large extent.

\begin{table}[h!]
\footnotesize
\centering
\captionsetup{justification=centering, font=small}
\setlength{\tabcolsep}{5.5pt}
\caption{R@1 scores speech $\to$ speech translation retrieval for various language pairs (VoxPopuli)}
\begin{tabular}{l@{\hskip 3pt}c@{\hskip 3pt}c@{\hskip 3pt}c@{\hskip 3pt}c@{\hskip 3pt}c@{\hskip 3pt}c@{\hskip 3pt}c@{\hskip 3pt}c}
\toprule
& \multicolumn{8}{c}{\textbf{X Speech $\to$ EN Speech retrieval}} \\
\toprule
\textbf{R@1} $\uparrow$  & \textbf{fr-en} & \textbf{pl-en} & \textbf{nl-en} & \textbf{es-en} & \textbf{hr-en} & \textbf{de-en} & \textbf{ro-en} & \textbf{cs-en} \\
\midrule
SONAR                & 91.91 & 95.79 & 95.16 & 95.30     & 52.16 & 94.18 & 94.45 & 95.62 \\
\singleSENSE\ & 96.55 & 96.46 & 95.75 & 96.48 & 96.5 & 94.71 & 
96.83 & 96.70 \\ \hline
\multiSENSE\  & 95.94 & 95.67 & 95.37 & 96.01 & 95.9 & 93.91 & 96.49 & 96.32 \\
\bottomrule
& \multicolumn{8}{c}{\textbf{EN Speech $\to$ Y Speech retrieval}} \\
\bottomrule
\textbf{R@1} $\uparrow$ & \textbf{en-fr} & \textbf{en-pl} & \textbf{en-nl} & \textbf{en-es} & \textbf{en-hr} & \textbf{en-de} & \textbf{en-ro} & \textbf{en-cs} \\
\midrule
SONAR            & 91.43 & 95.57 & 94.65 & 95.1     & 52.36 & 93.82 & 74.29 & 95.5  \\
\singleSENSE\ & 96.54 & 96.25 & 95.71 & 96.37 & 96.31 & 94.12 & 97.16 & 97.09 \\ \hline
\multiSENSE\  & 95.96 & 95.75 & 95.39 & 95.96 & 95.79 & 93.46 & 96.58 & 96.49 \\
\bottomrule
%\textbf{Size DB} & 49882 & 18768 & 10713 & 36320 & 7394 & 59116 & 16265 & 11181 \\
%\bottomrule
& \multicolumn{8}{c}{\textbf{X Speech $\to$ Y Speech retrieval}} \\
\bottomrule
\textbf{R@1} $\uparrow$ & \textbf{fr-de} & \textbf{hr-cs} & \textbf{ro-fr} & \textbf{hu-da} & \textbf{de-fr} & \textbf{cs-hr} & \textbf{fr-ro} & \textbf{da-hu} \\
\midrule
SONAR           & 91.39 & 52.93  &  92.01 &  3.32 & 92.73 & 53.65 & 92.15 & 4.31  \\
\singleSENSE\ & 95.20 & 94.75 & 96.55 & 92.69 & 95.36 & 94.69 & 96.90 & 92.63 \\ \hline
\multiSENSE\  & 93.83 & 94.01 & 96.18 & 91.07 & 93.72 & 93.91 & 96.49 & 90.79 \\
\midrule
%\textbf{Size DB} & 55656 & 4973 & 15745 & 3528 & 55656 & 4973 & 15745 & 3528 \\
%\midrule
\end{tabular}

\label{tab:Speech_speecH_retr}
\end{table}

Table~\ref{tab:mtedx_retr} and Table~\ref{tab:fleurs_retr} report R@1 scores for speech-to-text translation retrieval on MTEDx and FLEURS respectively. 
On MTEDx, the multi-task model remains close to \singleSENSE\ on most language pairs and stays above SONAR on the majority of them.
On FLEURS, our model stays close to \singleSENSE\ and consistently outperforms SONAR. 
In addition, the model slightly improves over \singleSENSE\ on the my-en pair (16.38 vs 14.11), suggesting that semantic generalization is preserved even in low-resource conditions.

\begin{table}[h!]
\footnotesize
\centering
\captionsetup{justification=centering, font=small}
\setlength{\tabcolsep}{5.5pt}
\renewcommand{\arraystretch}{0.92}
\caption{R@1 scores speech $\to$ text translation retrieval for various language pairs (MTEDx)}
\begin{tabular}{l@{\hskip 3pt}c@{\hskip 3pt}c@{\hskip 3pt}c@{\hskip 3pt}c@{\hskip 3pt}c@{\hskip 3pt}c@{\hskip 3pt}c@{\hskip 3pt}c@{\hskip 3pt}c}
\toprule
& \multicolumn{4}{c}{\textbf{X Speech $\to$ EN Text}} & \multicolumn{5}{c}{\textbf{X Speech $\to$ Y Text}} \\
\cmidrule(lr){2-5} \cmidrule(lr){6-10}
\textbf{R@1} $\uparrow$ & \textbf{it-en} & \textbf{fr-en} & \textbf{pt-en} & \textbf{ru-en} & \textbf{it-es} & \textbf{es-it} & \textbf{es-fr} & \textbf{fr-pt} & \textbf{pt-es} \\
\midrule
SONAR        & 89.01 & 82.45 & 85.05 & 84.76 & 92.25 & 88.57 & 87.76 & 84.27 & 87.08 \\
\singleSENSE\ & 90.69 & 87.01 & 86.69 & 83.06 & 94.35 & 86.83 & 87.35 & 90.08 & 89.31 \\ \hline
\multiSENSE\  & 90.10 & 86.14 & 85.68 & 82.27 & 94.15 & 86.57 & 86.91 & 89.68 & 88.63 \\
\bottomrule
\end{tabular}

\label{tab:mtedx_retr}
\end{table}

\vspace{-0.5cm}

\begin{table}[h!]
\footnotesize
\centering
\captionsetup{justification=centering, font=small}
\setlength{\tabcolsep}{5.5pt}
\renewcommand{\arraystretch}{0.92}
\caption{R@1 scores speech $\to$ text translation retrieval for various language pairs (FLEURS)}
\begin{tabular}{l@{\hskip 3pt}c@{\hskip 3pt}c@{\hskip 3pt}c@{\hskip 3pt}c@{\hskip 3pt}c@{\hskip 3pt}c@{\hskip 3pt}c@{\hskip 3pt}c}
\toprule
& \multicolumn{5}{c}{\textbf{X Speech $\to$ EN Text}} & \multicolumn{3}{c}{\textbf{X Speech $\to$ Y Text}} \\
\cmidrule(lr){2-6} \cmidrule(lr){7-9}
\textbf{R@1} $\uparrow$ & \textbf{ml-en} & \textbf{lb-en} & \textbf{uz-en} & \textbf{bs-en} & \textbf{my-en} & \textbf{ny-ces} & \textbf{sd-fr} & \textbf{xh-ar} \\
\midrule
SONAR &  22.36 & 39.98 & 54.12 & 27.27 & 8.19 & 18.18 & 21.20 & 16.07 \\
\singleSENSE\ & 62.59 & 60.88 & 55.40 & 60.39 & 14.11 & 27.00 & 59.74 & 36.67 \\ \hline
\multiSENSE\  & 61.55 & 59.59 & 55.17 & 60.28 & 16.38 & 25.24 & 58.24 & 35.71 \\
\bottomrule
\end{tabular}
\label{tab:fleurs_retr}
\end{table}

\vspace{-0.5cm}
\subsubsection{Speaker verification}

%Table~\ref{tab:speaker_verif} reports the speaker verification results on VoxCeleb1-O using Equal Error Rate (EER) and minimum Detection Cost Function (minDCF). 
%The results show that the multi-task model achieves an EER of 0.91\%, very close to the ECAPA-TDNN teacher (0.9\%), with a similarly small gap in minDCF. 
%This indicates that the speaker branch successfully learns to reproduce the teacher's speaker representations, even though the shared encoder is simultaneously optimized for semantic alignment. 
%More, this joint optimization seems to help the model to get a better speaker representation since the the \multiSENSE\ model outperforms \singleSENSEspk. 
%The proposed multi-task training thus preserves speaker-discriminative information effectively.

Table~\ref{tab:speaker_verif} reports the speaker verification results on VoxCeleb1-O using Equal Error Rate (EER) and minimum Detection Cost Function (minDCF). 
The results show that the multi-task model achieves an EER of 0.91\%, very close to the ECAPA-TDNN teacher (0.90\%), with a similarly small gap in minDCF. 
This indicates that the speaker branch successfully learns to reproduce the teacher's speaker representations, even though the shared encoder is simultaneously optimized for semantic alignment. 
Moreover, the joint optimization may even benefit speaker representations, as the \multiSENSE\ model slightly outperforms \singleSENSEspk.
Overall, the proposed multi-task training effectively preserves speaker-discriminative information.

\begin{table}[h!]
\footnotesize
\centering
\captionsetup{justification=centering, font=small}
\setlength{\tabcolsep}{7pt}
\renewcommand{\arraystretch}{0.92}
\caption{Speaker verification results on VoxCeleb1-O}
\begin{tabular}{l@{\hskip 3pt}c@{\hskip 3pt}c}
\toprule
\textbf{Model} & \textbf{EER} $\downarrow$ & $\mathbf{MinDCF_{0.01}}$ $\downarrow$ \\
\midrule
ECAPA-TDNN & 0.90 & 0.1104 \\

\singleSENSEspk    &  0.93 &  0.1285       \\ \hline
\multiSENSE\       & 0.91 &     0.1253  \\ 
\bottomrule
\end{tabular}

\label{tab:speaker_verif}
\end{table}
%where moved the speaker information?

\section{Analysis}

To better understand how the multi-task model exploits semantic and speaker information within the shared unified speech encoder, we analyze the learned layer-interpolation weights $\lambda_{\tau,\ell}$ (Eq.~\ref{eq:layer_weights}) of the two task-specific branches.

\begin{figure}[h!]
  \centering
  \includegraphics[width=0.9\linewidth]{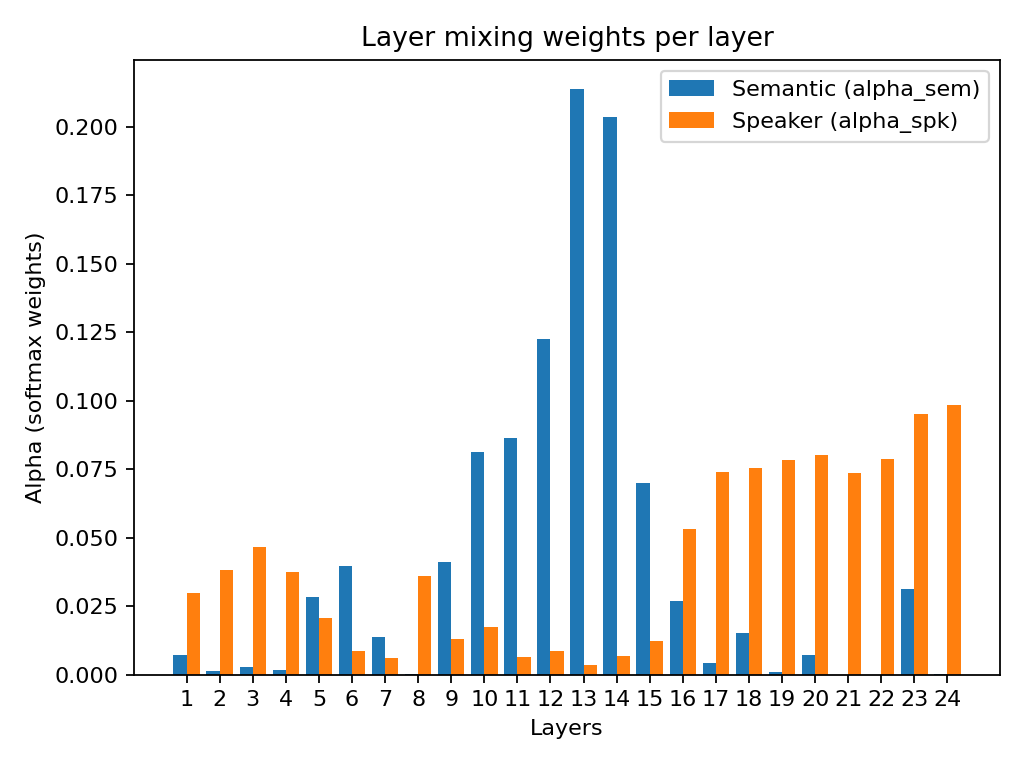}
  \caption{Learned layer-interpolation weights for the semantic and speaker branches of the unified speech encoder. }
  \label{fig:alpha_weights}
\end{figure}

Figure~\ref{fig:alpha_weights} shows the learned mixing weights. 
The two branches exhibit clearly different layer selection patterns. 
The semantic branch concentrates most of its weight on a narrow range of middle layers, with a strong peak around layers 13 and 14, indicating that the semantic task primarily relies on a localized region of the encoder.
In contrast, the speaker branch distributes its weights more broadly across the entire encoder, with a gradual increase toward the highest layers, peaking at layers 23 and 24, indicating that the speaker task draws on a wider portion of the network.

%Interestingly, the two branches show a complementary pattern: in the layers where one branch assigns high weights, the other assigns very small or near-zero weights. For instance, around layers 13 and 14 where the semantic weights peak, the speaker weights are very small, and conversely, in the upper layers where the speaker branch assigns its highest weights, the semantic branch contributes very little. This shows that the model automatically learns to select different encoder layers for each task.

\section{Conclusion}

In this work, we introduced a unified post-training framework that enables a single speech foundation model to learn multiple utterance-level attribute representations simultaneously through task-specific branches attached to a shared encoder.

Our experiments show that multiple attributes such as semantic and speaker representations can be jointly learned without major degradation of either task: the multi-task model remains close to the single-task semantic baseline on both speech-to-speech and speech-to-text retrieval, including on low-resource languages, while speaker verification performance stays nearly on par with the ECAPA-TDNN teacher. The analysis of learned layer-interpolation weights reveals that each task selects different encoder layers with a complementary pattern, showing how the model automatically learns to select the most useful layers for each task.

In future work, we plan to extend this framework by incorporating additional attributes such as emotion, language, and accent, toward building richer and more versatile speech representations from a unified speech encoder.

\bibliographystyle{IEEEtran}
\bibliography{mybib}

\end{document}